\documentclass[letterpaper]{article}
\usepackage[preprint]{aaai2027}
\usepackage[hyphens]{url}
\usepackage{graphicx}
\usepackage{natbib}
\usepackage{caption}
\usepackage{algorithm}
\usepackage{algorithmic}
\usepackage{booktabs}
\usepackage{amsmath}
\usepackage{amssymb}
\usepackage{booktabs}
\usepackage{tabularx}
\usepackage{array}
\usepackage{multirow}

\title{LAVE: Latent Visual Evidence-Enhanced Planning for Video Tool-use Agents}
\author{
Zijian Wang\equalcontrib\textsuperscript{\rm 1},
Junnan Zhu\equalcontrib\textsuperscript{\rm 2},
Rongzhen Li\equalcontrib\textsuperscript{\rm 3},
Xiao Liu\equalcontrib\textsuperscript{\rm 1},
Guohui Xiang\textsuperscript{\rm 3},\\
Quan Lu\textsuperscript{\rm 3},
Lijia Liu\textsuperscript{\rm 1},
Yining Wang\textsuperscript{\rm 4},
Jiang Zhong\textsuperscript{\rm 1}\corresponding,
Kaiwen Wei\textsuperscript{\rm 1}\corresponding
}
\affiliations{
\footnotesize
\textsuperscript{\rm 1}College of Computer Science, Chongqing University, Chongqing, China\\
\textsuperscript{\rm 2}MAIS, Institute of Automation, Chinese Academy of Sciences\\
\textsuperscript{\rm 3}Chongqing National Data AI Research Institute, AI Research Lab
\quad \textsuperscript{\rm 4}Unisound AI Technology Co.Ltd\\
zjstud@cqu.edu.cn, weikaiwen@cqu.edu.cn
}
\begin{document}
\maketitle

\begin{abstract}
Long-video understanding requires models to efficiently acquire and reuse
sparse visual evidence from long and redundant video streams. Recent video
tool-use agents address this challenge by iteratively invoking visual Tools at
different temporal scales, but their Tool--Planner communication typically
relies on textual observations. Such text-only interfaces provide lossy
summaries of Tool computations, causing previously computed visual evidence
not verbalized to be discarded and unavailable for subsequent planning. We
identify this limitation as the \textit{Tool observation bottleneck} and propose
\textbf{LA}tent \textbf{V}isual \textbf{E}vidence-Enhanced Planning
(\textbf{LAVE}), a training-free framework for reusing latent visual evidence
from completed Tool calls. LAVE introduces a dual-channel observation
interface: the visible channel preserves the original textual trajectory, while
the latent channel stores pre-verbal visual updates with their Tool roles,
source-frame timestamps, and visual locations. During planning, LAVE retrieves
evidence relevant to the current Planner state but not covered by textual
observations, and integrates it through bounded timestamp-aligned latent
updates with entropy-constrained frame-time routing. This enables video agents
to reuse existing visual computation without additional training, frame replay,
or modifications to the original orchestration. Extensive experiments on
Video-MME, LongVideoBench, and CG-Bench show that LAVE consistently improves
video tool-use agents across backbones. Under a comparable frame budget, LAVE
improves the Video-MME overall score by 3.76 points over the strongest baseline,
demonstrating the effectiveness of latent visual evidence reuse for multi-step video-agent planning.

\end{abstract}

\section{Introduction}

Long-video understanding requires locating sparse answer-relevant evidence from long and redundant visual streams~\cite{tang2026videounderstanding,nguyen2024videolanguage,meng2026watchrememberreasonhumanview}. Processing all frames is costly and often introduces irrelevant information. Recent MLLMs have thus moved beyond the conventional ``watch-then-answer'' paradigm toward agentic video understanding, where a Planner iteratively reasons about the query, invokes visual Tools at different temporal scales, and uses their observations to guide subsequent search~\cite{videoseek2026,framemind2025,videotir2026,sage2025,videomind2025}. This paradigm enables adaptive video understanding by allocating visual computation based on the current planning context.


\begin{figure}[t]
\centering
\includegraphics[width=\columnwidth]{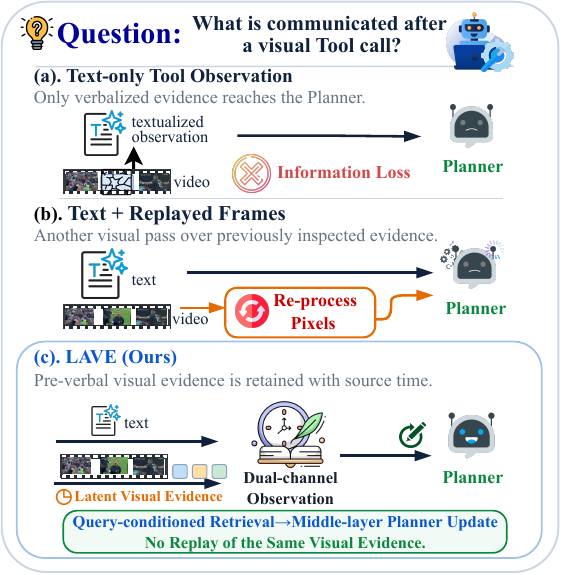}
\caption{Comparison of video agent Tool--Planner communication paradigms. Text-only observations discard unverbalized visual evidence, while raw visual inputs require additional visual inference. LAVE preserves the text trajectory and reuses temporally grounded latent visual evidence from completed Tool calls to improve subsequent planning.
} 
\label{fig:protocol-comparison}
\end{figure}


Recent video agents have largely focused on improving how visual evidence is acquired and organized. Methods such as VideoSeek~\cite{videoseek2026} and FrameMind~\cite{framemind2025} use hierarchical operations, including Overview, Skim, and Focus, to move from coarse exploration to local inspection. However, after a Tool processes the selected frames, its output is usually compressed into a short textual observation, and the Planner makes subsequent decisions from this text. This interface creates a less studied problem. The agent must preserve acquired visual evidence for future planning, rather than only acquire it.

We call this problem the \textit{Tool observation bottleneck}. As shown in
Figure~\ref{fig:protocol-comparison}(a), visual evidence can enter the reasoning
context only after being converted into a textual observation. This
verbalization is inherently lossy and may also be misaligned with the
Planner's actual information needs. For example, when the next decision depends
on an action occurring in the scene, the observation may instead emphasize the
number or identities of visible people, thereby omitting task-relevant evidence
while introducing irrelevant details. Once the Tool call ends, the discarded
visual signals are no longer available for subsequent planning.
An alternative is to replay the frames to a
vision-language Planner, as illustrated in
Figure~\ref{fig:protocol-comparison}(b). However, the Planner must then
re-encode and reinterpret the same pixels at every reasoning step. Moreover,
text and visual inputs differ substantially in information density, abstraction
level, and noise, while how the Planner balances these modalities remains
implicit; it may therefore rely disproportionately on the more compact textual
context. Repeated visual processing also introduces additional inference
latency and computational cost.

To meet these requirements, we propose \textbf{LA}tent \textbf{V}isual \textbf{E}vidence Enhanced Planning (\textbf{LAVE}), a training-free framework that reuses visual evidence from completed Tool calls during subsequent planning. LAVE introduces a dual-channel observation interface: the visible channel preserves the original textual trajectory, while the latent channel stores pre-verbal visual updates from the Tool forward pass. Each latent entry is associated with its Tool role, source frame time, and visual location, preserving both visual content and temporal provenance. At each planning step, LAVE retrieves evidence relevant to the current Planner state but not covered by the textual observation, adapts temporal retrieval coverage based on entropy, and integrates selected evidence through bounded residual updates at corresponding timestamped observations. This enables the Planner to reuse omitted visual information for future Tool selection without modifying the visible trajectory, additional training, or frame replay. The latent channel is disabled during final answer generation, affecting answers only through improved planning and evidence acquisition.

Experiments on Video-MME~\cite{videomme2025}, LongVideoBench~\cite{wu2024longvideobench}, and CG-Bench~\cite{chen2025cg} benchmarks demonstrate that LAVE effectively enhances existing video tool-use agents through a training-free and plug-in inference-time framework. LAVE requires no parameter updates, additional training, or modifications to the original agent orchestration. When applied to the open-source Qwen3.5-9B VLM, LAVE improves Video-MME performance from 48.50\% to 62.77\%, yielding a 14.27-point gain over the original VLM baseline. Furthermore, when integrated into the Qwen3.5-9B-based VideoSeek~\cite{videoseek2026} agent, LAVE further improves the VideoSeek baseline from 59.01\% to 62.77\% with a comparable frame budget, showing that LAVE reuses previously computed visual evidence without increasing the frame budget. In summary, the contributions of this work are as follows:

\begin{itemize}
    \item We identify the Tool observation bottleneck in video tool-use agents, where text-only Tool observations discard previously computed valuable visual evidence that could benefit subsequent planning.

    \item We propose LAVE, a training-free dual-channel interface that preserves time-anchored latent visual evidence from completed Tool calls and reuses it through query-aware, entropy-constrained frame-time routing and bounded Planner updates.

    \item Experiments on three benchmarks and two model backbones show that LAVE consistently improves video agent planning under comparable frame budgets, including a 3.76 point gain on Video-MME, without additional training or repeated visual inference.
\end{itemize}

\section{Related Work}

\subsection{Video Question Answering}
Existing video question answering (Video QA) methods mainly focus on constructing effective visual
evidence for the reasoning model. VideoTree~\citep{videotree2025}, BOLT~\citep{bolt2025},
Q-Frame~\citep{qframe2025}, and GIFT~\citep{gift2026} select or organize
query-relevant frames, while TSPO~\citep{tspo2026} and A.I.R.~\citep{air2026}
optimize temporal sampling. Other methods, such as Video-EM~\citep{videoem2025}
and ReFineVQA~\citep{refinevqa2026}, represent evidence through event memories
or refined descriptions, while Video-R1~\citep{videor12025} improves reasoning
over visual inputs. Despite different designs, these methods share an
evidence-centric interface, where visual information is prepared before being
consumed by the reasoning model. 
Recent agentic video QA systems incorporate evidence acquisition into an iterative reasoning process. VideoSeek~\citep{videoseek2026} and FrameMind~\citep{framemind2025} enable targeted visual inspection through adaptive tool interactions, while VideoTIR~\citep{videotir2026}, SAGE~\citep{sage2025}, and VideoMind~\citep{videomind2025} explore multi-step planning and agent coordination. By dynamically deciding what visual evidence to acquire at each step, these systems move beyond fixed-frame processing toward active video understanding. However, these methods focus on acquiring relevant evidence, but visual information omitted from explicit observations cannot be reused for subsequent planning. In this paper, we propose LAVE, which addresses this limitation in the Tool--Planner interface by preserving temporally grounded latent visual evidence for inference-time reuse.



\begin{figure*}[!t]
\centering
\includegraphics[width=0.98\textwidth]{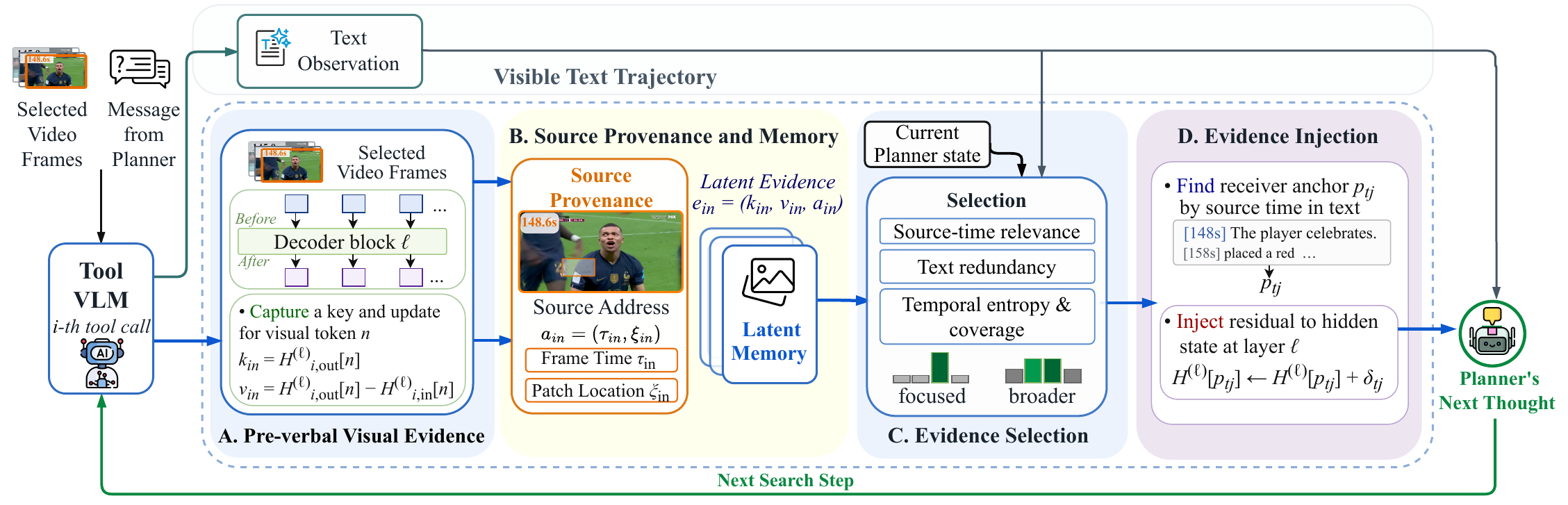}
\caption{Overview of LAVE.
Selected frames and Planner messages are processed by the Tool VLM, producing
text observations and pre-verbal visual states. During Tool execution, LAVE
(A) extracts latent key--value evidence and (B) stores addressed evidence with
source provenance. During the next Planner prefill, LAVE (C) retrieves relevant
source-time groups using the current Planner state and (D) injects bounded
residuals at resolved receiver anchors to guide the next thought and Tool
action.}
\label{fig:lave-method}
\end{figure*}

\subsection{Latent Communication in Agents}
Recent work explores continuous representations as a complement to textual
communication. Communicating Activations~\citep{activationcomm2025},
LatentMAS~\citep{latentmas2026}, and State Delta Encoding
(SDE)~\citep{statedelta2025} transfer intermediate activations, latent states,
or token-aligned state changes across agents. Activation
Addition~\citep{turner2023activation} further shows that additive hidden-state
interventions can steer subsequent computation. In multimodal models,
DeepStack~\citep{deepstack2024} introduces visual representations through
intermediate residual paths, Vision Function Layer~\citep{vfl2025} identifies
layer-specific visual functions, and Latent Visual
Reasoning~\citep{lvr2026} preserves visual semantics without fully
verbalizing them. However, these methods mainly communicate agent-level states
or enrich the current inference process, rather than preserving visual
evidence omitted during verbalization for later decisions. LAVE instead
retains pre-verbal visual block updates alongside the explicit textual
trajectory, allowing useful evidence to remain available for subsequent
planning.

\section{Method}
\label{sec:method}

\paragraph{Task Definition and Overview.}
This work focuses on agentic video QA, which requires models to iteratively
acquire and reason over visual evidence through interactions with specialized
Tools. Specifically, given a question, Planner selects appropriate Tools to
inspect different video regions, and the resulting observations guide
subsequent reasoning and tool selection. LAVE augments this iterative
framework with a dual-channel observation mechanism: the original textual
observation is preserved as the visible trajectory, while pre-verbal visual
evidence produced during Tool execution is retained through a latent channel.

Figure~\ref{fig:lave-method} illustrates the overall framework of LAVE. The
latent channel captures intermediate visual states before text generation,
associates them with source-frame timestamps, and stores them for retrieval. At each planning step, LAVE retrieves evidence that is relevant to current decision but not covered by textual observation, and integrates
it to Planner through a timestamp-aligned latent update. Temporal entropy
adaptively controls the temporal coverage of retrieval, while a utility-based
filter prevents uninformative evidence from influencing Planner. The latent
channel is only activated during planning and is disabled during final generation.

\paragraph{Pre-verbal Visual Evidence Acquisition.}
LAVE retains visual information from the selected video frames before the Tool
VLM verbalizes it, allowing omitted evidence to remain available for later
planning. Figure~\ref{fig:lave-method}(A) shows the selected frames and Planner
message entering the Tool VLM. Let \(C_i\) denote this \(i\)-th Tool call, and
let \(n\) index the visual tokens produced from its selected frames.

At decoder block \(\ell\),
\(H_{i,\mathrm{in}}^{(\ell)}[n]\) and
\(H_{i,\mathrm{out}}^{(\ell)}[n]\) denote the hidden states of visual token
\(n\) immediately before and after the block, respectively. In \(k_{in}\) and
\(v_{in}\), the paired subscript \(in\) combines Tool-call index \(i\) and
visual-token index \(n\). The roman labels \(\mathrm{in}\) and
\(\mathrm{out}\) on \(H\) correspond to the states before and after block
\(\ell\). LAVE constructs:
\begin{equation}
\begin{aligned}
k_{in}
&= H_{i,\mathrm{out}}^{(\ell)}[n],\\
v_{in}
&= H_{i,\mathrm{out}}^{(\ell)}[n]
- H_{i,\mathrm{in}}^{(\ell)}[n].
\end{aligned}
\label{eq:tool-side-key-value}
\end{equation}


The state after block \(\ell\) serves as the key because it represents the
visual content available for later relevance matching. The difference between
the after- and before-block states serves as the value because it isolates the
visual update introduced by the block. Using the complete hidden state as the
value would also retransmit information already present before the block and
could unnecessarily perturb the Planner. The shared visual-token index links
each retrieval key to its corresponding visual update.

\paragraph{Source Provenance and Memory.}
The latent update alone describes visual content but does not identify the
video moment that produced it.  Figure~\ref{fig:lave-method}(B) associates the
key--value evidence from the Pre-verbal Visual Evidence Acquisition stage with its source
video time and visual location before storing it in latent memory.

Each frame carries its video timestamp, while visual preprocessing
retains the frame index and patch coordinate associated with every visual
token. When multiple selected frames are jointly encoded, the frame index
preserves the token-to-frame correspondence. For visual token \(n\) from Tool
call \(i\), this correspondence provides the source-frame time
\(\tau_{in}\) and within-frame visual location \(\xi_{in}\). As above, the
paired subscript \(in\) denotes Tool-call index \(i\) and visual-token index
\(n\). LAVE forms the source address:
\begin{equation}
a_{in}=(\tau_{in},\xi_{in}).
\end{equation}

The frame time supports temporal grouping and subsequent alignment, while the
visual location preserves the spatial origin of the token.

The Source Provenance and Memory stage attaches this source address to the key
and visual update captured during the Pre-verbal Visual Evidence Acquisition
stage, forming the latent evidence
\begin{equation}
e_{in}=(k_{in},v_{in},a_{in}),
\label{eq:addressed-visual-evidence}
\end{equation}
here, \(k_{in}\) supports relevance matching, \(v_{in}\) carries the
pre-verbal visual update, and \(a_{in}\) records where the evidence originated.
Keeping the three fields linked helps distinguish visually similar evidence
observed at different video moments and enables source-time routing without
discarding spatial provenance. 
For example, Figure~\ref{fig:lave-method}(B) shows visual evidence originating
at 148.6\,s. The Tool may verbalize ``The player celebrates,'' while the
associated visual updates retain jersey identity, gesture, and nearby ball
motion. LAVE stores these updates together with the 148.6\,s source time,
allowing a later player-identification decision to retrieve omitted evidence
from the relevant moment.

\begin{figure}[t]
\centering
\includegraphics[width=\columnwidth]{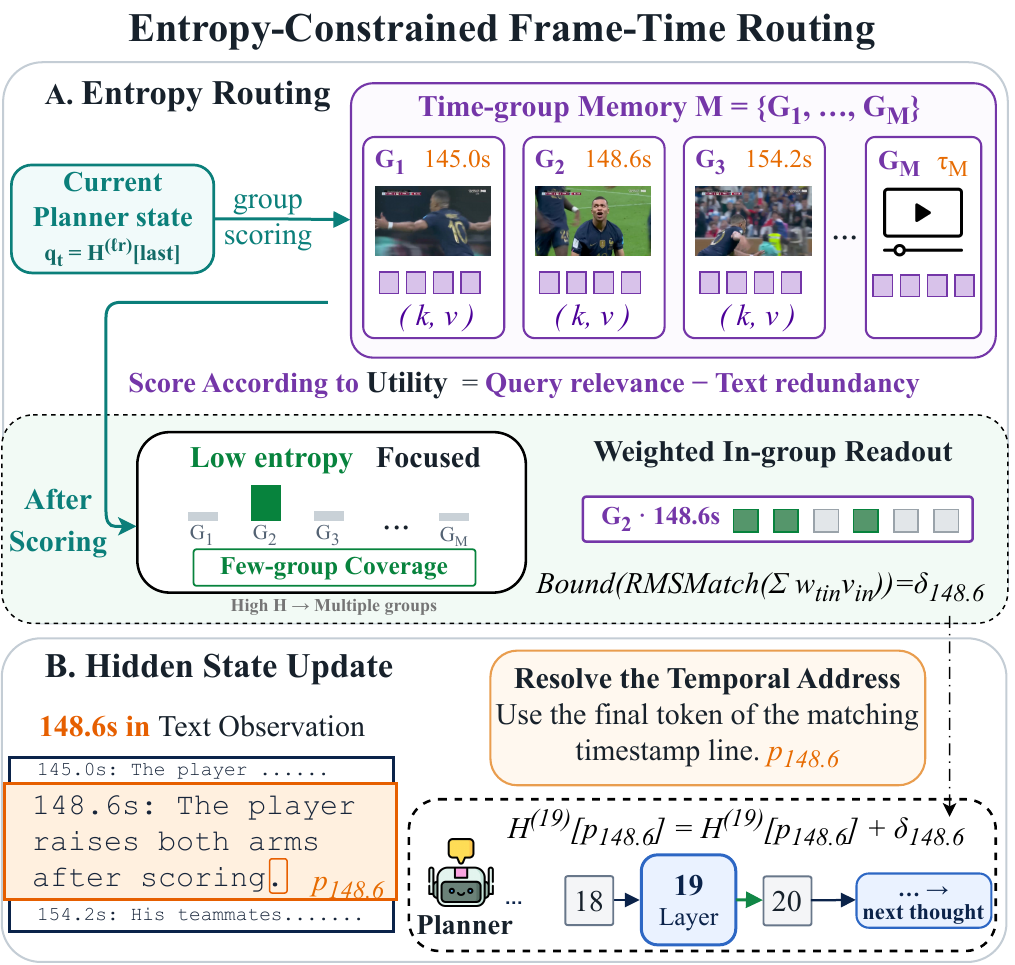}
\caption{Entropy-constrained frame-time routing.
(A) Planner-state relevance and temporal entropy select source-time groups
and form bounded residuals.
(B) Each residual is written at the timestamp-matched receiver anchor during
Planner prefill.}
\label{fig:frame-time-routing}
\end{figure}

\paragraph{Evidence Selection.}
Figure~\ref{fig:lave-method}(C) summarizes evidence selection from latent
memory using the current Planner state and visible text trajectory. Source-time
relevance identifies useful evidence, text redundancy discounts already
verbalized content, and temporal entropy adapts retrieval from focused to broad
coverage. These components determine relevance, complementarity, and temporal
coverage, respectively.

Figure~\ref{fig:frame-time-routing}(A) details how selection operates over
time-group memory. At planning step \(t\), let \(L_t\) denote the length of the
current Planner prompt. LAVE uses the final-token state after decoder block
\(\ell\) as the current Planner state:
\begin{equation}
q_t=H_{t,\mathrm{out}}^{(\ell)}[L_t]\in\mathbb{R}^{d}.
\label{eq:planner-query}
\end{equation}

This state summarizes the question, visible trajectory, and current planning
context. The Planner state and Tool-side keys are extracted at the same decoder
block, allowing cosine similarity to compare them directly. For latent entry
\(e_{in}\), source-time relevance is
\(r_{tin}=\cos(q_t,k_{in})\). A high relevance score indicates that the visual
content represented by \(k_{in}\) is useful for the current planning state.

Relevance alone may retrieve content already expressed in the visible
trajectory. Let \(\mathcal O_i^{(\ell)}\) contain the block-\(\ell\) states of
the textual observation generated by Tool call \(C_i\). LAVE assigns visual
token \(n\) the utility
\begin{equation}
u_{tin}=r_{tin}-\lambda
\left[\max_{o\in\mathcal O_i^{(\ell)}}\cos(k_{in},o)\right]_+,
\label{eq:evidence-utility}
\end{equation}
where \([x]_+=\max(x,0)\). The two terms measure query relevance and textual
redundancy, respectively, so high utility identifies useful but insufficiently
verbalized evidence. In the shown example, jersey or gesture cues receive high
utility when they support player identification beyond ``The player
celebrates.''

Source times partition latent memory into groups \(G_1,\ldots,G_M\), each
containing the keys and visual updates from one video moment. LAVE aggregates
the strongest token utilities within each group into its score \(s_{tj}\).
Group-first scoring treats multiple high-scoring patches from the same frame as
one temporal candidate, avoiding redundant token-level allocation.

A temperature-scaled softmax over the group scores yields a source-time
relevance distribution \(p_t=(p_{t1},\ldots,p_{tM})\), where \(p_{tj}\)
denotes the relevance assigned to group \(G_j\). A concentrated distribution
indicates that one source time dominates the current decision, whereas a
diffuse distribution indicates that several moments remain plausible. LAVE
uses the Shannon entropy~\cite{shannon1948mathematical} \(\mathrm H(p_t)\) to match temporal coverage to this
concentration:
\begin{equation}
\mathcal J_t^\star
=
\underset{\varnothing\neq\mathcal J\subseteq[M]}{\arg\max}\;
\sum_{j\in\mathcal J}p_{tj}
\quad\mathrm{s.t.}\quad
\log|\mathcal J|\leq\mathrm H(p_t).
\label{eq:entropy-matched-coverage}
\end{equation}

The objective retains source-time groups with the largest relevance mass, while
the entropy constraint adaptively determines the number of retained groups.
When the relevance distribution is concentrated, low entropy leads to
\emph{Focused Coverage}, where retrieval concentrates on a few dominant
source-time groups. Conversely, when multiple source times have comparable
relevance, higher entropy leads to \emph{Broad Coverage}, which preserves
evidence from multiple plausible moments. Unlike fixed-width retrieval, this
design adapts temporal coverage to the current Planner state while preserving
the utility-based ordering within each group. In the example of
Figure~\ref{fig:frame-time-routing}(A), \(G_2\) at 148.6\,s dominates the
relevance distribution, resulting in Focused Coverage. If the groups at
145.0\,s, 148.6\,s, and 154.2\,s receive similar scores, the higher entropy
instead expands retrieval to Broad Coverage.
After entropy determines the retained source-time groups, LAVE removes entries
with non-positive utility and reads each group independently. For every
retained group \(G_j\), the remaining utilities determine the token weights,
group readout, and bounded residual:
\begin{equation}
\begin{aligned}
w_{tin}
&=
\underset{\substack{e_{in}\in G_j\\u_{tin}>0}}
{\operatorname{softmax}}\!\left(u_{tin}/\tau_v\right),\\
\bar v_{tj}
&=
\sum_{\substack{e_{in}\in G_j\\u_{tin}>0}}
w_{tin}v_{in},\\
\delta_{tj}
&=
\operatorname{Bound}\!\left(
\alpha_{c_i}
\frac{\operatorname{RMS}(q_t)}
{\operatorname{RMS}(\bar v_{tj})}
\bar v_{tj}
\right).
\end{aligned}
\label{eq:bounded-group-readout}
\end{equation}
Here, \(w_{tin}\) is the within-group utility weight, \(\tau_v\) is the
softmax temperature for token-level readout, and \(\alpha_{c_i}\) is the gain
associated with the role \(c_i\) of source Tool call \(C_i\). The weighted
readout \(\bar v_{tj}\) combines complementary visual updates from the same moment. Root-mean-square (RMS)~\cite{burden2015numerical} matching aligns its scale with the current Planner state,
while \(\operatorname{Bound}(\cdot)\) limits the accumulated latent change.
Each resulting residual \(\delta_{tj}\) retains the source time of \(G_j\) for
subsequent injection.
Detailed derivations are provided in Appendix~\ref{app:method-derivations}.

\begingroup
\newcommand{\MainResultSep}{\vrule width 0.20pt}
\newcommand{\MainResultText}[1]{#1}
\newcommand{\MainResultBest}[1]{\textbf{#1}}
\newcommand{\MainResultGroup}[1]{%
  \specialrule{0.45pt}{0pt}{0pt}
  \multicolumn{10}{@{}l@{}}{%
    \rule{0pt}{2.45ex}\hspace{3pt}\textbf{\textit{#1}}%
  }\\[-0.05ex]
  \specialrule{0.45pt}{0pt}{0pt}
}

\begin{table*}[t!]
  \centering
  \fontsize{8.2pt}{8.8pt}\selectfont
  \setlength{\tabcolsep}{1.8pt}
  \renewcommand{\arraystretch}{1.1}

  \begin{tabular}{
    @{}
  >{\raggedright\arraybackslash}p{6.70cm}
  @{\hspace{7pt}\MainResultSep\hspace{1pt}}
  >{\centering\arraybackslash}p{0.85cm}
  >{\centering\arraybackslash}p{1.03cm}
  >{\centering\arraybackslash}p{0.82cm}
  >{\centering\arraybackslash}p{0.94cm}
  >{\centering\arraybackslash}p{0.90cm}
  @{\hspace{4pt}\MainResultSep\hspace{4pt}}
  >{\centering\arraybackslash}p{0.94cm}
  >{\centering\arraybackslash}p{0.90cm}
  @{\hspace{4pt}\MainResultSep\hspace{4pt}}
  >{\centering\arraybackslash}p{0.94cm}
  >{\centering\arraybackslash}p{0.90cm}
  @{}
  }
    \toprule
    \multicolumn{1}{l}{%
      \multirow{2}{*}{\raisebox{-0.65ex}{\textbf{Model + Method}}}}
      & \multicolumn{5}{c}{\textbf{Video-MME}}
      & \multicolumn{2}{c}{\textbf{LongVideoBench}}
      & \multicolumn{2}{c}{\textbf{CG-Bench}} \\
    \cmidrule(lr){2-6}
    \cmidrule(lr){7-8}
    \cmidrule(l){9-10}
    \multicolumn{1}{l@{\hspace{7pt}\MainResultSep\hspace{1pt}}}{}
  & Short$\uparrow$ & Medium$\uparrow$ & Long$\uparrow$ & Overall$\uparrow$ & Frames$\downarrow$
  & Overall$\uparrow$ & Frames$\downarrow$ & Overall$\uparrow$ & Frames$\downarrow$ \\

    \MainResultGroup{Closed-source or proprietary-API systems}
    GPT-4o~\citep{gpt4o2024}
      & 80.0 & 70.3 & 65.3 & 71.9 & 256
      & 66.7 & 256 & 45.2 & 256 \\
    Gemini-1.5-Pro~\citep{gemini152024}
      & 81.7 & 74.3 & 67.4 & 75.0 & 256
      & 64.0 & 256 & 37.2 & 256 \\
    GPT-5~\citep{gpt52025}
      & -- & -- & 67.9 & -- & 384
      & 64.5 & 384 & -- & -- \\
    GPT-4 + VideoTree~\citep{videotree2025}
      & -- & -- & -- & 54.2 & 128
      & -- & -- & -- & -- \\
    Gemini-2.5-Flash + SAGE-Flash~\citep{sage2025}
      & -- & -- & -- & 63.5 & 128+
      & -- & -- & -- & -- \\

    \MainResultGroup{Open-source systems with method-specific training}
    Qwen2.5-VL-7B + VideoTIR$^\dagger$~\citep{videotir2026}
      & 67.2 & 55.8 & 50.7 & 57.9 & 16+
      & 53.1 & 16+ & -- & -- \\
    Qwen2.5-VL-7B + Video-R1$^\dagger$~\citep{videor12025}
      & -- & -- & 50.2 & 59.3 & 32
      & 56.4 & 32 & 34.4 & 32 \\
    Qwen2.5-VL-7B + FrameMind$^\dagger$~\citep{framemind2025}
      & 66.0 & 64.8 & 61.2 & 64.0 & 64+
      & -- & -- & -- & -- \\
    Qwen2-VL-7B + VideoMind$^\dagger$~\citep{videomind2025}
      & -- & -- & 49.2 & 58.2 & 64+
      & 56.3 & 64+ & 38.4 & 64+ \\
    Qwen3-VL-8B + MACF$^\dagger$~\citep{macf2026}
      & -- & -- & -- & 60.4 & 96
      & 56.8 & 96 & -- & -- \\
    Qwen2.5-VL-7B + TSPO$^\dagger$~\citep{tspo2026}
      & -- & -- & 56.4 & 65.5 & 1fps
      & 62.1 & 1fps & -- & -- \\

    \MainResultGroup{Open-source systems without method-specific training}
    Qwen2.5-VL-7B + Video-EM$^\ddagger$~\citep{videoem2025}
      & 72.4 & 60.3 & 53.4 & 62.0 & 1fps
      & 59.6 & 1fps & 38.1 & 1fps \\
    Qwen2-VL-7B + BOLT$^\ddagger$~\citep{bolt2025}
      & 69.4 & 57.5 & 51.5 & 59.5 & 1fps
      & -- & -- & -- & -- \\
    Qwen2-VL-7B + Q-Frame$^\ddagger$~\citep{qframe2025}
      & 69.4 & 57.1 & 48.3 & 58.3 & 128
      & 58.4 & 128 & 36.7 & 128 \\
    Qwen2.5-VL-7B + GIFT$^\ddagger$~\citep{gift2026}
      & 76.8 & 63.2 & 53.1 & 64.4 & 128
      & 61.3 & 128 & 38.2 & 128 \\
    LLaVA-OV-7B + ReFineVQA$^\ddagger$~\citep{refinevqa2026}
      & -- & -- & -- & 63.1 & 200
      & -- & -- & -- & -- \\
    Qwen2.5-VL-7B + A.I.R.$^\ddagger$~\citep{air2026}
      & -- & -- & -- & 65.0 & 1fps
      & 61.4 & 1fps & 39.1 & 200 \\

    \specialrule{0.75pt}{0pt}{0pt}
    Qwen3.5-9B CoT$^\ddagger$~\citep{latentvc2026}
      & -- & -- & -- & 48.5 & 64
      & -- & -- & -- & -- \\
    GLM-4.6V-Flash-9B + VideoSeek$^\ddagger$
      & 72.07 & 55.24 & 47.83 & 58.38 & 64.08
      & 55.12 & 72.36 & 33.80 & 74.14 \\
    GLM-4.6V-Flash-9B + LAVE$^\ddagger$ (Ours)
      & 73.50 & 58.18 & 49.43 & 60.37 & 62.02
      & 56.52 & 72.02 & 34.12 & 74.02 \\
    Qwen3.5-9B + VideoSeek$^\ddagger$
      & 70.78 & 55.95 & 48.21 & 59.01 & 60.70
      & 55.69 & \MainResultBest{68.22}
      & 34.65 & 71.29 \\
    \textbf{Qwen3.5-9B + LAVE$^\ddagger$ (Ours)}
      & \MainResultBest{76.92}
      & \MainResultBest{58.33}
      & \MainResultBest{52.98}
      & \MainResultBest{62.77}
      & \MainResultBest{60.66}
      & \MainResultBest{58.38}
      & 69.80
      & \MainResultBest{38.61}
      & \MainResultBest{71.03} \\

    \bottomrule
  \end{tabular}
  \caption{Experiment results on different benchmarks without subtitles.
$\dagger$: methods requiring additional training or method-specific optimization;
$\ddagger$: inference-only methods without additional training. \textbf{Bold} numbers indicate the best inference-only performance on open-sourced MLLMs.}
\label{tab:main-results}
\end{table*}
\endgroup

\paragraph{Evidence Injection.}
Figure~\ref{fig:lave-method}(D) summarizes receiver-side writing. Each retained
source-time group provides a bounded residual \(\delta_{tj}\) from
Eq.~\ref{eq:bounded-group-readout} together with its source time. During
Planner prefill, LAVE resolves a receiver anchor in the visible text trajectory
and writes the residual at receiver block \(\ell_r\). In our implementation,
the receiver and evidence extraction blocks are identical, i.e.,
\(\ell_r=\ell\). The remaining Planner blocks then propagate the updated state
to the next thought and Tool action.

Figure~\ref{fig:frame-time-routing}(B) illustrates this process for the
source-time group at 148.6\,s. The source time first identifies the
corresponding timestamped line, ``148.6\,s: The player raises both arms after
scoring.'' LAVE uses the final token of this line as receiver anchor
\(p_{tj}\) and writes \(\delta_{148.6}\) to its hidden state. The source time
and receiver anchor therefore serve different roles: the former identifies the
video moment, whereas the latter specifies where the associated residual enters
the Planner sequence.

The line-final position is used because its hidden state summarizes the
timestamp and the complete textual observation on that line. Moreover, the
position precedes the subsequent causal states, allowing later Planner tokens
to jointly use the visible description and its aligned pre-verbal visual
evidence. The operation changes only the hidden state; the visible observation
and its tokenization remain unchanged.

Each selected group retains an independent residual and receiver anchor.
Residuals whose source times resolve to the same receiver position are
accumulated:
\begin{equation}
\widetilde H_t^{(\ell_r)}[p]
=
H_t^{(\ell_r)}[p]
+
\sum_{\substack{j\in\mathcal J_t^\star\\p_{tj}=p}}
\delta_{tj}.
\label{eq:addressed-residual-injection}
\end{equation}
Here, \(H_t^{(\ell_r)}[p]\) is the original Planner state at receiver position
\(p\), and the summation contains only residuals assigned to that position.
Consequently, evidence from different source times remains separately
addressed unless the corresponding timestamped observations share the same
receiver anchor.
\paragraph{Tool--Planner Interaction Flow.}
Tool capture and Planner writing occur in consecutive agent turns. A completed
Tool prefill stores addressed evidence; during the next Planner prefill, LAVE
uses the block-\(\ell\) prompt states to route residuals and update their
receiver anchors via Eq.~\ref{eq:addressed-residual-injection}. Subsequent
blocks integrate the timestamped text and aligned latent evidence before the
next Tool action.

\flushbottom

\section{Experiments}


\paragraph{Benchmarks.} 
We evaluate LAVE on the full Video-MME benchmark~\citep{videomme2025}, the LongVideoBench validation set~\citep{wu2024longvideobench}, and the CG-Bench mini set~\citep{chen2025cg}. All benchmarks are evaluated without subtitles. Video-MME measures multiple-choice understanding over short, medium, and long videos, while LongVideoBench and CG-Bench provide complementary long-context and temporal-grounding evaluations. We report each benchmark's overall score and sampled frames per question.

\paragraph{Baselines.} 
We compare LAVE with three groups of systems: (1) closed-source or
proprietary-API systems, including GPT-4o~\citep{gpt4o2024},
Gemini-1.5-Pro~\citep{gemini152024}, GPT-5~\citep{gpt52025}, and
tool-augmented systems such as VideoSeek~\citep{videoseek2026} and
SAGE-Flash~\citep{sage2025}; (2) open-source systems with method-specific
training, including VideoTIR~\citep{videotir2026}, FrameMind~\citep{framemind2025},
VideoMind~\citep{videomind2025}, MACF~\citep{macf2026}, and TSPO~\citep{tspo2026};
and (3) open-source inference-only systems without additional training,
including Video-EM~\citep{videoem2025}, BOLT~\citep{bolt2025},
GIFT~\citep{gift2026}, ReFineVQA~\citep{refinevqa2026}, and A.I.R.~\citep{air2026}.
We use VideoSeek~\citep{videoseek2026} with its Overview--Skim--Focus loop as
the primary agent baseline.

\paragraph{Implementation Details.}

Our controlled comparison evaluates
Qwen3.5-9B~\citep{qwen35model2026} and
GLM-4.6V-Flash-9B~\citep{glm46vflash2025}. For each backbone, LAVE and the
no-latent reference use identical configurations and differ only in whether
stored latent evidence is routed into planning steps; final answering remains
text-only.
The main configuration uses aligned capture and receiver blocks
\(\ell=\ell_r=19\), up to eight retrieved evidence tokens per planning step,
a \(1.0\times\) residual gain, and entropy-constrained temporal coverage.
The key hyperparameters are ablated below; all remaining settings and
system details are reported in Appendix~\ref{app:experimental-settings}.



\begin{figure}[t]
\centering
\includegraphics[width=\columnwidth]
{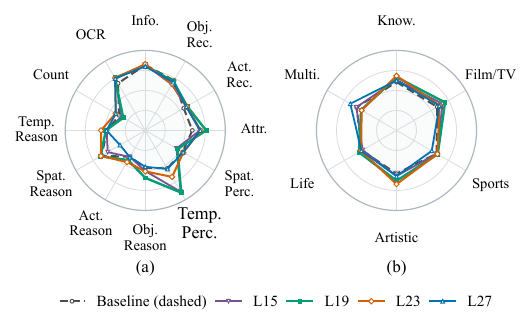}
\caption{Effects of latent-evidence depth across blocks. (a) task-level
results; (b) content-level results.}
\label{fig:ablation-summary}
\end{figure}

\begin{table}[t]
\centering
\small
\setlength{\tabcolsep}{4.5pt}
\begin{tabular}{@{}lrrrr@{}}
\toprule
Variant & Overall & Short & Medium & Long \\
\midrule
Gaussian noise & 49.28 & 60.87 & 43.48 & 43.48 \\
No latent & 59.01 & 70.78 & 55.95 & 48.21 \\
\midrule
Block 15 & 59.08 & 69.31 & 58.42 & 49.50 \\
Block 19 & \textbf{61.39} & 70.30 & \textbf{62.38} & \textbf{51.49} \\
Block 23 & 61.06 & \textbf{71.29} & 61.39 & 50.50 \\
Block 27 & 57.43 & 69.31 & 59.41 & 43.56 \\
\bottomrule
\end{tabular}
\caption{{Ablation on Communication-block selection.}
We disable entropy-constrained frame-time routing and evaluate different blocks
for Tool-side evidence extraction and Planner-side residual writing.
}
\label{tab:latent-injection-ablation}
\end{table}

\paragraph{Main Experiment.}
Table~\ref{tab:main-results} reports results across Video-MME,
LongVideoBench, and CG-Bench. We find: (1) Among inference-only systems with a
comparable frame budget, LAVE achieves the strongest performance. On
Qwen3.5-9B, LAVE improves Video-MME from 48.50\% to 62.77\% (+14.27 points)
and surpasses VideoSeek by 3.76 points with nearly identical frames
(60.66 vs.\ 60.70). It also generalizes across backbones, improving
GLM-4.6V-Flash-9B + VideoSeek from 58.38\% to 60.37\% with fewer frames.
(2) Compared with closed-source and training-based systems, LAVE achieves a
better performance-efficiency trade-off. While these systems often rely on
larger visual budgets or additional optimization, LAVE reaches competitive
performance with open-source backbones under a controlled inference budget.
(3) Duration-wise analysis shows consistent gains of 4.14, 2.38, and 4.77
points on short, medium, and long videos, respectively. The largest gain on
long videos indicates that latent evidence reuse is especially beneficial when
information spans multiple Tool interactions. (4) Interestingly, LAVE improves
over VideoSeek with a comparable but slightly smaller frame budget (60.66 vs.\
60.70), suggesting that latent evidence reuse enables more effective Tool
exploration trajectories rather than simply increasing visual processing.
(5) Beyond Video-MME, LAVE improves LongVideoBench by 2.69 points and CG-Bench
by 3.96 points, demonstrating consistent gains across benchmarks.
\paragraph{Ablation Study on Communication-block Selection.}
To investigate the communication block selection, following existing work~\citep{statedelta2025,glimpseprune2025}, we evaluate full-attention blocks around
\(2L/3\). As shown in Table~\ref{tab:latent-injection-ablation}, Block 19
achieves the best overall performance, with particularly strong improvements on
medium and long videos. Replacing latent evidence with Gaussian noise causes a
substantial performance drop, confirming that the gains come from structured
visual evidence rather than hidden-state perturbations. Figure~\ref{fig:ablation-summary}
further shows that different blocks capture complementary task-specific
information, with block 19 providing stronger temporal perception.




\begin{table}[t!]
\centering

\small
\setlength{\tabcolsep}{3pt}
\begin{tabular}{@{}lccr@{}}
\toprule
Setting & Blocks & Retrieval & Overall \\
\midrule
No latent & -- & -- & 59.01 \\
+ Evidence & 19 & Flat token & 61.39 \\
+ Routing & 19 & Group$\to\mathcal H\to$token & 62.77 \\
\bottomrule
\end{tabular}
\caption{Incremental contributions of latent visual evidence and
frame-time routing.}
\label{tab:incremental-ablation}
\end{table}

\paragraph{Ablation Study on Latent Evidence Routing.}
Using block 19 as the evidence carrier, we evaluate the contribution of
entropy-constrained frame-time routing. As shown in
Table~\ref{tab:incremental-ablation}, replacing flat token retrieval with
adaptive routing improves accuracy from 61.39\% to 62.77\%, demonstrating the
benefit of source-time-aware evidence selection. Table~\ref{tab:hyperparam-sensitivity}
further studies three factors: retrieved evidence tokens, residual gain, and
temporal coverage. We vary only the target factor while keeping Block 19 as the evidence carrier, other routing
settings, planning-only injection, and evaluation protocol unchanged. Across
token budgets \(4/8/16\), gains \(0.5/1.0/2.0\times\), and coverage of
\(2/4/8\) groups, intermediate settings perform best, indicating a balance
between insufficient evidence and excessive interference. Entropy-adaptive
coverage consistently outperforms fixed group counts, validating dynamic
temporal selection based on the Planner state.



\begin{table}[t!]
\centering
\small
\setlength{\tabcolsep}{4pt}
\renewcommand{\arraystretch}{0.85}
\begin{tabular}{@{}llr@{}}
\toprule
Factor & Setting & Accuracy \\
\midrule
\multirow{3}{*}{Evidence tokens}
& 4  & 60.26 \\
& 8  & \textbf{62.77} \\
& 16 & 59.01 \\
\midrule
\multirow{3}{*}{Residual gain}
& \(0.5\times\) & 60.26 \\
& \(1.0\times\) & \textbf{62.77} \\
& \(2.0\times\) & 60.58 \\
\midrule
\multirow{4}{*}{Temporal coverage}
& Fixed 2 groups & 59.57 \\
& Fixed 4 groups & 61.32 \\
& Fixed 8 groups & 60.48 \\
& Entropy-adaptive & \textbf{62.77} \\
\bottomrule
\end{tabular}
\caption{Sensitivity of latent readout and temporal coverage.}
\label{tab:hyperparam-sensitivity}
\end{table}

\paragraph{Planner--Tool Behavior Analysis.}
\begin{table}[t!]
\centering
\small
\setlength{\tabcolsep}{3.4pt}
\begin{tabular}{@{}lrrrrr@{}}
\toprule
Method & Calls/QA & Hit@0 & Hit@3 & Hit@5 & Hit@10 \\
\midrule
No latent & 6.139 & 65.63 & 76.69 & 79.01 & 84.11 \\
LAVE & {6.132} & {72.78} & {79.09}
& {81.62} & {86.41} \\
\midrule
\(\Delta\) & \(-0.007\) & \(+7.15\) & \(+2.41\) & \(+2.62\) & \(+2.30\) \\
\bottomrule
\end{tabular}
\caption{Video-level clue coverage on CG-Bench. }
\label{tab:video-clue-coverage}
\end{table}
CG-Bench provides answer-bearing clue times, enabling direct evaluation of
LAVE's impact on Planner--Tool temporal exploration. Hit@\(r\%\) measures
whether a Skim or Focus interval overlaps with a ground-truth clue expanded by
\(r\%\) of the video duration. As shown in
Table~\ref{tab:video-clue-coverage}, LAVE keeps Tool calls nearly unchanged
while improving exact clue overlap by 7.15 points. The smaller gains under
larger tolerances indicate that the baseline often searches near relevant
regions but misses precise localization. By preserving omitted visual evidence
with source-time provenance, LAVE guides subsequent Tool calls toward more
accurate regions without additional exploration.

\begin{figure}[t]
\centering
\includegraphics[width=\columnwidth]{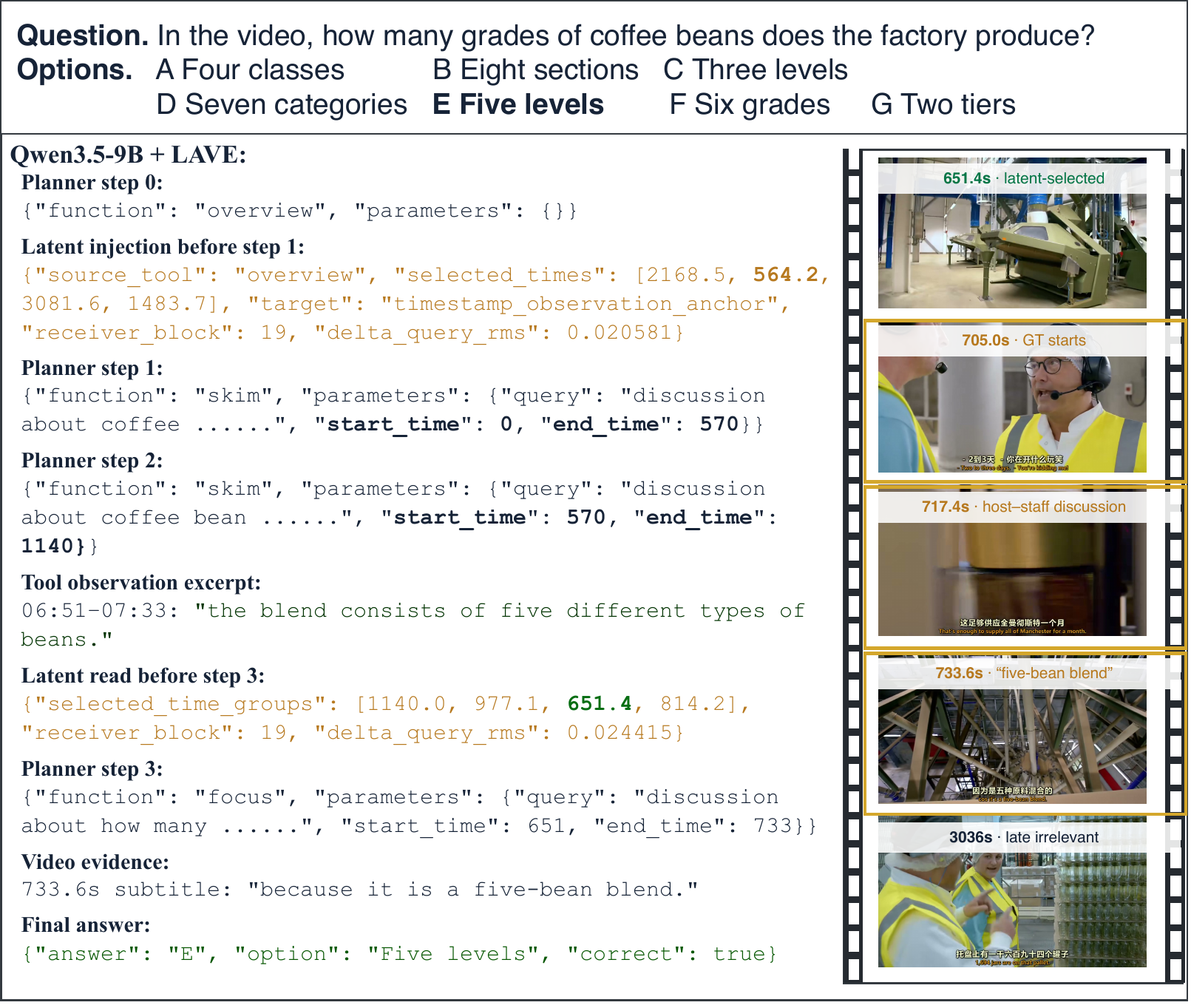}
\caption{Qualitative example of LAVE.}
\label{fig:q7317-case}
\end{figure}

\paragraph{Case Study.}

Figure~\ref{fig:q7317-case} illustrates how LAVE improves temporal
exploration. After identical initial Tool calls, the second Skim observation
reveals a five-bean blend, while latent routing further retrieves evidence
around 651.4\,s and guides the Planner to issue Focus over 651--734\,s. This
interval overlaps the answer-bearing segment and captures the key evidence near
733.6\,s, leading to the correct answer E. The case shows that LAVE
improves planning by refining the Tool exploration trajectory through reuse of
previously acquired visual evidence rather than directly providing the answer.
\section{Conclusion}
This paper identifies the \textit{Tool observation bottleneck} of video tool-use agents: textual Tool--Planner communication discards visual evidence omitted during verbalization. To solve this problem, we propose LAVE, a training-free framework that preserves this evidence in a time-anchored latent channel and reuses it through entropy-constrained routing. Experiments on Video-MME, LongVideoBench, and CG-Bench show consistent gains across backbones, including a 3.76-point improvement in Video-MME overall score under a comparable frame budget. These results demonstrate the effectiveness of latent visual evidence reuse for multi-step video-agent planning.


\bibliography{references}

\appendix
\setcounter{secnumdepth}{2}
\setcounter{subsection}{0}
\renewcommand{\thesubsection}{\Alph{subsection}}
\section*{Appendix}
\label{app:method-details}

Section~A provides the formal derivations referenced by the Method section.
Section~B reports the evaluation protocol, numerical settings, runtime, and
backbone details referenced by the Experiments section. Sections~C and D
present the paired case study and limitations.

\subsection{Method Derivations}
\label{app:method-derivations}

This subsection expands the four Method operations mathematically. Figure~2(A)
and (B) map to pre-verbal evidence and source provenance below. Figure~2(C)
maps to utility scoring and entropy-constrained routing. Figure~2(D) maps to
timestamp-aligned bounded residual injection. Concrete run values are
centralized in Table~\ref{tab:routing-config}.

\subsubsection{Notation and Tensor Conventions}

All sequence positions are one-based unless a processor coordinate is
explicitly described as zero-based. For a positive integer $m$, we write
$[m]=\{1,\ldots,m\}$ and
$[m]_0=\{0,\ldots,m-1\}$. Boldface is not used for vectors; whether a symbol
is a scalar, vector, set, or sequence is stated below and again when it first
appears in an equation. The superscript $(\ell)$ always identifies a decoder
block, whereas a parenthesized superscript such as $(h)$ on a utility denotes
an order statistic rather than a network layer.

\begin{center}
\centering
\fontsize{9}{10.2}\selectfont
\setlength{\tabcolsep}{2.0pt}
\renewcommand{\arraystretch}{1.04}
\begin{tabular}{>{\raggedright\arraybackslash}p{0.13\columnwidth}
                >{\raggedright\arraybackslash}p{0.20\columnwidth}
                >{\raggedright\arraybackslash}p{0.59\columnwidth}}
\toprule
Symbol & Domain & Definition \\
\midrule
$i$ & $[I]$ & Tool call; $I$ is the call count. \\
$t$ & $[T]$ & Planner turn; $T$ is the turn count. \\
$n$ & $[L_i]$ & Tool-prefill position; $n\in\mathcal V_i$ is visual. \\
$j$ & $[M]$ & Source-time group; $M$ groups. \\
$p$ & $[L_t]$ & Prompt position; length $L_t$. \\
$h$ & Local & Rank, coordinate, or summation index. \\
$z$ & $[N]$ & Evaluation item; $N$ questions. \\
$\ell,\ell_r$ & $\mathbb N_0$ & Capture / receiver block. \\
$d$ & $\mathbb N$ & Hidden width; vectors lie in $\mathbb R^d$. \\
$c_i$ & Role & Overview, Skim, or Focus. \\
\bottomrule
\end{tabular}
\captionof{table}{Index notation for Appendix A.}
\label{tab:app-index-notation}
\end{center}

For a Tool prefill, $H_{i,\mathrm{in}}^{(\ell)}$ and
$H_{i,\mathrm{out}}^{(\ell)}$ are matrices in
$\mathbb R^{L_i\times d}$ immediately before and after block $\ell$; indexing
with $[n]$ selects one row in $\mathbb R^d$. For a Planner prefill, the same
notation with turn index $t$ gives matrices in
$\mathbb R^{L_t\times d}$. The tuple $e_{in}$ is a stored evidence record, not
a vector to which arithmetic is directly applied. Calligraphic symbols denote
sets except $\mathcal H_t$, which denotes Shannon entropy. Scalar similarity,
utility, probability, temperature, gain, and norm variables are written in
lower case; vector updates use $v$, $\delta$, $D$, or $\Delta$.

\subsubsection{Pre-verbal Evidence and Source Provenance}
\label{app:evidence-construction}

\paragraph{Prefill-only visual positions.}
Let $C_i$ denote the $i$-th Tool call and let
$X_i=(x_{i1},\ldots,x_{iL_i})$ be its processed multimodal prefill sequence.
Here $L_i\in\mathbb N$ is the sequence length and $x_{ip}$ is the processed
input element at position $p\in[L_i]$, which may originate from text or a
visual patch. Let $m_{in}\in\{0,1\}$ be the processor-provided indicator that
position $n$ is visual. The candidate-position set is
\begin{equation}
\mathcal V_i=\{n\in[L_i]:m_{in}=1\}.
\label{eq:app-visual-set}
\end{equation}
Thus $\mathcal V_i\subseteq[L_i]$ contains exactly the visual rows from which
latent evidence may be captured, and $|\mathcal V_i|$ is the uncompressed
candidate count for call $i$. The value $m_{in}=1$ includes a visual position;
$m_{in}=0$ excludes it.
The mask excludes system instructions, the question, timestamp text, and all
ordinary text positions. Evidence is captured by hooks during the original
Tool generation prefill. The hooks are removed after this prefill, so states
created while generating the textual observation are never stored. Thus the
latent channel contains pre-verbal contextual visual states rather than hidden
states of the generated answer.

\paragraph{Block key and block-update value.}
At decoder block $\ell$, the implementation records the block input and output
at every $n\in\mathcal V_i$:
\begin{equation}
\begin{aligned}
k_{in}&=H_{i,\mathrm{out}}^{(\ell)}[n],\\
v_{in}&=H_{i,\mathrm{out}}^{(\ell)}[n]
       -H_{i,\mathrm{in}}^{(\ell)}[n].
\end{aligned}
\label{eq:app-tool-kv}
\end{equation}
Both $k_{in}$ and $v_{in}$ are vectors in $\mathbb R^d$. The key $k_{in}$ is
the contextual visual state at position $n$ after block $\ell$, and the value
$v_{in}$ is the same position's block-local update. The paired subscript $in$
always means ``Tool call $i$, visual position $n$''; it is not a product.
A residual decoder block can be written abstractly as
\begin{equation}
H_{i,\mathrm{out}}^{(\ell)}
=H_{i,\mathrm{in}}^{(\ell)}
+F_{\ell}\!\left(H_{i,\mathrm{in}}^{(\ell)}\right).
\label{eq:app-residual-block}
\end{equation}
In Eq.~\eqref{eq:app-residual-block},
$F_\ell:\mathbb R^{L_i\times d}\rightarrow
\mathbb R^{L_i\times d}$ denotes the complete residual branch implemented by
decoder block $\ell$, including the transformations whose net result is added
to the incoming state. The equation is an interface-level identity and does
not assume that attention and feed-forward sublayers are a single operation.
Therefore $v_{in}=F_{\ell}(H_{i,\mathrm{in}}^{(\ell)})[n]$ is exactly the net
update introduced by block $\ell$ at that visual position. Using the block
output as the key preserves a content-rich address for matching, while using
the difference as the value avoids retransmitting the entire incoming state.

Block identifiers in the paper and released configuration are zero-based. In
particular, ``Block 19'' is
\texttt{language\_model.layers[19]}. The implementation obtains its input and
output from pre- and post-forward hooks on that block; it does not index a
\texttt{hidden\_states} tuple. Consequently, the extra embedding entry often
present at \texttt{hidden\_states[0]} cannot introduce an off-by-one error.

\paragraph{Exact time and spatial addresses.}
Let $\mathcal F_i$ be the finite set of source units shown in call $i$; a
source unit is an input frame for Skim/Focus or a montage tile for Overview.
The map $f_i:\mathcal V_i\rightarrow\mathcal F_i$ identifies the source unit
that generated visual position $n$. The timestamp map
$T_i:\mathcal F_i\rightarrow\mathbb R_{\geq0}$ returns that unit's time in
seconds. The address stored with the evidence is
\begin{equation}
a_{in}=\left(\tau_{in},\xi_{in}\right),\qquad
\tau_{in}=T_i(f_i(n)),
\label{eq:app-source-address}
\end{equation}
Here $\tau_{in}\in\mathbb R_{\geq0}$ is the verified source timestamp and
$\xi_{in}$ is a discrete spatial coordinate identifying the patch inside its
source frame or montage tile. Therefore
$a_{in}=(\tau_{in},\xi_{in})$ is metadata rather than a hidden vector; it is
used for grouping and receiver resolution and is never added to a model state.
The mapping is
Tool-specific because Overview and the two local Tools use different visual
layouts.

For \textbf{Overview}, every montage image has a $2\times4$ tile layout. Let
$U_i\geq1$ be the number of temporal montage slices produced for call $i$,
$u\in[U_i]_0$ the zero-based slice index, $r\in\{0,1\}$ the tile row, and
$c\in\{0,1,2,3\}$ the tile column. The flattened zero-based tile index is
\begin{equation}
b(u,r,c)=8u+4r+c.
\label{eq:app-montage-index}
\end{equation}
Consequently $b(u,r,c)\in[8U_i]_0$ uniquely identifies one tile among all
slices, and $T_i[b(u,r,c)]$ is the timestamp printed for that tile. The symbols
$r$ and $c$ are local montage coordinates and are unrelated to relevance
$r_{tin}$ or Tool role $c_i$.
The timestamp matrix placed immediately before the montage image provides
$T_i[b(u,r,c)]$. Processor grid coordinates assign each visual patch to one
tile, and every retained patch in that tile receives the corresponding
timestamp. Hence a montage token is never treated as an unaddressed element in
a flat 64-token sequence. The tile partition organizes provenance but does not
reduce the candidate set.

For \textbf{Skim} and \textbf{Focus}, processor coordinates retain the image
index, temporal index, patch row, and patch column of each visual position.
The image and temporal indices select the exact input-frame timestamp, while
the row and column form $\xi_{in}$. Equivalently, $\xi_{in}$ is the tuple of
processor coordinates needed to recover the within-frame patch, while
$f_i(n)$ uses its image and temporal components to recover the source frame.
Therefore the three Tools share the same
memory schema even though Overview is montage-based and Skim/Focus operate on
individual sampled frames.

The latent evidence defined in the main text is the triplet
\begin{equation}
e_{in}=\bigl(k_{in},v_{in},a_{in}\bigr),
\label{eq:app-memory-entry}
\end{equation}
whose fields are matching key $k_{in}\in\mathbb R^d$, transferable update
$v_{in}\in\mathbb R^d$, and source address $a_{in}$. The implementation also
stores side metadata $\mu_{in}=(c_i,i)$, where $c_i$ is the Tool role and $i$
is the origin-call identifier. Equivalently, an implementation record is
$\widetilde e_{in}=(e_{in},\mu_{in})$, but $\mu_{in}$ is not part of the
transferred vector or the main-text definition of $e_{in}$. Memory is scoped
to the current question and cleared before the next question.

\subsubsection{Complementarity-aware Evidence Utility}
\label{app:utility-derivation}

\paragraph{Planner query in the shared block space.}
At planning turn $t$, let $L_t\in\mathbb N$ be both the token length and final
one-based position of the rendered Planner prompt. The query is the output of
the same block used for Tool-side capture:
\begin{equation}
q_t=H_{t,\mathrm{out}}^{(\ell)}[L_t].
\label{eq:app-query}
\end{equation}
Here $H_{t,\mathrm{out}}^{(\ell)}\in\mathbb R^{L_t\times d}$ is the Planner
state after block $\ell$, and $q_t\in\mathbb R^d$ is its final-token row. The
subscript $t$ indicates a planning turn rather than a video timestamp. Because
causal attention allows the final token to read all preceding prompt tokens,
$q_t$ summarizes the question and visible trajectory available at this turn.
Because $q_t$ and $k_{in}$ are outputs of the same backbone block, they have a
shared dimensional coordinate space and can be compared without a learned
projection or adapter. This permits a train-free similarity computation. It
does not guarantee semantic alignment between same-block states. The
communication-block ablation evaluates this design choice indirectly through
downstream performance. For numerical stability, define
$\widehat{x}=x/\max(\lVert x\rVert_2,\epsilon)$ with
$\epsilon=10^{-6}$. Query relevance is
\begin{equation}
r_{tin}=\widehat q_t^{\top}\widehat k_{in}.
\label{eq:app-relevance}
\end{equation}
For any $x\in\mathbb R^d$, $\widehat x$ is its safeguarded unit-normalized
version, $\lVert\cdot\rVert_2$ is the Euclidean norm, and
$\epsilon>0$ prevents division by zero. Thus $r_{tin}\in[-1,1]$ is the cosine
similarity between the turn-$t$ query and the key from call $i$, position $n$.

\paragraph{Text-redundancy penalty.}
Let $\mathcal P_i$ be the token positions occupied by the visible observation
returned by call $C_i$. The positions are resolved from tokenizer character
offsets; exact token-subsequence matching is used only when offsets are not
available. At the same Planner prefill and block $\ell$, define
\begin{equation}
\mathcal O_i^{(\ell)}
=\{H_{t,\mathrm{out}}^{(\ell)}[p]:p\in\mathcal P_i\}.
\label{eq:app-observation-states}
\end{equation}
$\mathcal P_i\subseteq[L_t]$ is therefore a set of Planner prompt positions,
and $\mathcal O_i^{(\ell)}\subset\mathbb R^d$ is the corresponding set of
block-output text states. The symbol $p$ indexes a prompt position, whereas
$o\in\mathcal O_i^{(\ell)}$ denotes one text-state vector.
The amount of key content already represented in text is
\begin{equation}
d_{tin}=
\left[
\max_{o\in\mathcal O_i^{(\ell)}}
\widehat k_{in}^{\top}\widehat o
\right]_+,
\label{eq:app-redundancy}
\end{equation}
where $[\zeta]_+=\max(\zeta,0)$ is the positive-part operator for scalar
$\zeta\in\mathbb R$. Consequently
$d_{tin}\in[0,1]$ is the largest nonnegative cosine similarity between key
$k_{in}$ and any visible observation state from the same Tool call. The
resulting utility is
\begin{equation}
u_{tin}=r_{tin}-\lambda d_{tin}.
\label{eq:app-utility}
\end{equation}
$u_{tin}\in\mathbb R$ is a scalar complementarity score, and
$\lambda\geq0$ controls how strongly text-covered content is discounted. As
specified in the main text, group scoring
and entropy routing use these utilities before sign filtering; entries with
$u_{tin}\leq0$ are removed only after source-time groups have been selected and
before value readout. If a Tool observation cannot be
resolved exactly in the current prompt, its associated evidence is skipped;
the implementation does not approximate the missing text span with the whole
prompt.

The within-forward ordering is strictly read-then-write. All reported
configurations use $\ell_r=\ell$. This block first
produces an unmodified $H_{t,\mathrm{out}}^{(\ell)}$; LAVE reads $q_t$ and all
$\mathcal O_i^{(\ell)}$ from that state, computes retrieval and residuals, and
only then modifies the resolved rows passed to block $\ell+1$. Therefore the
query and redundancy states used at turn $t$ cannot contain the latent update
being computed at the same turn, avoiding a circular dependency.

This utility separates two questions that plain query--key similarity
conflates. The first term asks whether the latent visual state is useful for the
current planning decision. The second asks whether the same content has already
been conveyed by the Tool text. High utility therefore favors relevant visual
information that is complementary to the visible observation.

\subsubsection{Entropy-constrained Frame-time Routing}
\label{app:entropy-routing}

\paragraph{Source-time group score.}
After timestamp verification, entries from one Tool call with the same
normalized source timestamp form a source-time group. The implementation uses
the pair (origin call, normalized timestamp) as the group identity. This
instantiates the main-text notion that each group represents one video moment
while retaining Tool-role and call provenance; groups from distinct calls may
therefore remain distinct even when their timestamps are numerically equal,
and their residuals may later share a receiver anchor.

At Planner turn $t$, let $G_1,\ldots,G_M$ be all timestamp-valid groups
available before sign filtering. Every $G_j$ has one origin call $i(j)$ and one
verified source time $\tau_j$, and
$G_j=\{e_{i(j)n}:\tau_{i(j)n}=\tau_j\}$. The number $M$ is turn-dependent
through the available trajectory, but the turn subscript is omitted to match
the main text. The map $i(j)$ is used wherever the group's Tool role or
call-specific observation is needed.

Let $\mathcal U_{tj}$ be the indexed collection
\(\{u_{t,i(j),n}:e_{i(j)n}\in G_j\}\), and order its scalar utilities as
$u_{tj}^{(1)}\geq\cdots\geq u_{tj}^{(|\mathcal U_{tj}|)}$. The group score is
\begin{equation}
s_{tj}=\frac{1}{m_j}\sum_{h=1}^{m_j}u_{tj}^{(h)},
\qquad
m_j=\min(4,|\mathcal U_{tj}|).
\label{eq:app-group-score}
\end{equation}
Here $u_{tj}^{(h)}$ is the $h$-th largest utility in group $j$ at turn $t$,
$m_j\in\{1,2,3,4\}$ is the number averaged, and
$s_{tj}\in\mathbb R$ is the resulting group-level score. The rank index
$h$ has no relation to hidden width coordinate indices used later.
Using the mean of the strongest four responses lets several
supporting patches raise a frame-time candidate without allowing a frame with
many weak patches to dominate merely because it contains more tokens.

The source-time distribution is
\begin{equation}
p_{tj}=\frac{\exp(s_{tj}/\tau_g)}
{\sum_{h=1}^{M}\exp(s_{th}/\tau_g)},
\qquad
\mathcal H_t=-\sum_{j=1}^{M}p_{tj}\log p_{tj},
\label{eq:app-group-distribution}
\end{equation}
Here $\tau_g>0$ is the group-softmax temperature, $p_{tj}\in(0,1)$ is the
probability assigned to source-time group $j$, and
$p_t=(p_{t1},\ldots,p_{tM})$ satisfies $\sum_jp_{tj}=1$.
$\mathcal H_t\in[0,\log M]$ is the Shannon entropy in nats because the natural
logarithm is used. If $M=0$, the
distribution and entropy are not evaluated and the latent update for turn $t$
is the zero update.

\paragraph{Closed-form solution of the coverage constraint.}
The entropy constraint is a routing design choice rather than a consequence
that is uniquely implied by information theory. Once this formulation is
chosen, however, its optimal subset and cardinality have the exact solution
derived below.
The main text selects the largest probability mass under
$\log|\mathcal J|\leq\mathcal H_t$, where
$\mathcal J\subseteq[M]$ is a nonempty candidate subset of source-time groups
and $|\mathcal J|$ is its cardinality. For a fixed integer cardinality $K$, an exchange
argument shows that the maximizing subset contains the $K$ largest
probabilities: replacing any selected $p_{ta}$ by an unselected
$p_{tb}>p_{ta}$ strictly increases the objective. Because all probabilities
are positive, the objective is nondecreasing in $K$. The largest feasible
cardinality is therefore
\begin{equation}
K_t=\max\!\left(1,
\left\lfloor\exp(\mathcal H_t)\right\rfloor\right).
\label{eq:app-entropy-cardinality}
\end{equation}
In the exchange argument, $a,b\in[M]$ are generic group indices.
In this expression, $\exp(\mathcal H_t)$ is the entropy-induced effective
group count, $\lfloor\cdot\rfloor$ is the floor operator, and $K_t\geq1$ is the
largest integer cardinality allowed by the entropy constraint before resource
limits are applied. With $M$ groups and integer total evidence-token
budget $B\geq1$, the implemented
cardinality is
\begin{equation}
\widetilde K_t=\min(M,B,K_t),\qquad
\mathcal J_t^\star=\operatorname{TopK}(p_t,\widetilde K_t).
\label{eq:app-entropy-solution}
\end{equation}
When $M>0$, $\widetilde K_t$ is an integer satisfying
$1\leq\widetilde K_t\leq\min(M,B)$ and is the feasible number of retained groups.
$\operatorname{TopK}(p_t,k)$ returns the indices of the $k$ largest entries of
$p_t$, and $\mathcal J_t^\star\subseteq[M]$ is the selected group-index set.
The integer argument $k$ of $\operatorname{TopK}$ is a requested cardinality
and is unrelated to evidence key vectors $k_{in}$.
Ties are resolved by the deterministic order of the stored groups and do not
alter the budget. When $M=0$, we define
$\widetilde K_t=0$ and $\mathcal J_t^\star=\varnothing$.
Thus $\exp(\mathcal H_t)$ acts as the effective number of plausible source
times. A peaked distribution approaches one retained group, whereas a diffuse
distribution permits broader temporal coverage.

\paragraph{Global token-budget allocation.}
After entropy routing, non-positive entries are removed exactly as stated in the
main text. Define the readable retained groups as
\begin{equation}
\mathcal J_t^+=\left\{j\in\mathcal J_t^\star:
\exists\,e_{i(j)n}\in G_j\ \text{with}\ u_{t,i(j),n}>0\right\}.
\label{eq:app-positive-groups}
\end{equation}
Thus $\mathcal J_t^+\subseteq\mathcal J_t^\star$ contains the selected groups
that still have at least one readable entry. If $\mathcal J_t^+=\varnothing$,
the turn produces no latent write. Otherwise, selecting groups does not imply
reading every positive visual token in them. For each
$j\in\mathcal J_t^+$, let
$n_j^{\star}\in\mathcal V_{i(j)}$ be the visual-position index of the
highest-utility entry in selected group $G_j$:
\begin{equation}
n_j^\star=\underset{n:\,e_{i(j)n}\in G_j,\ u_{t,i(j),n}>0}{\arg\max}\;
u_{t,i(j),n}.
\label{eq:app-best-group-entry}
\end{equation}
The allocator first reserves one token for every selected group:
\begin{equation}
\mathcal S_{tj}^{(0)}=\{n_j^{\star}\},
\qquad j\in\mathcal J_t^+.
\label{eq:app-mandatory-token}
\end{equation}
Thus $\mathcal S_{tj}^{(0)}$ is the singleton set containing the mandatory
visual-position index for group $j$; superscript $(0)$ denotes allocation
initialization, not a decoder block.
The remaining $B-|\mathcal J_t^+|$ slots are filled by the globally largest
utilities among all unselected positive entries in the retained groups. This
two-stage allocation guarantees temporal coverage first and then spends the
remaining capacity on the strongest patch-level evidence. The final sets
$\mathcal S_{tj}\subseteq\mathcal V_{i(j)}$ contain the visual-position
indices allocated to group $j$ and satisfy
\begin{equation}
\sum_{j\in\mathcal J_t^+}|\mathcal S_{tj}|
\leq B,
\qquad
|\mathcal S_{tj}|\geq1.
\label{eq:app-budget-guarantee}
\end{equation}
The first inequality is a global token budget across all retained times; the
second guarantees at least one selected visual update per retained group.

\subsubsection{Timestamp-aligned Bounded Residual Injection}
\label{app:bounded-injection}

\paragraph{Within-group value readout.}
For each readable retained group $j\in\mathcal J_t^+$, only the allocated
positive-utility entries in $\mathcal S_{tj}$ enter the value readout. For the
remainder of this paragraph, $i=i(j)$ denotes the unique origin call of group
$j$. The normalized scalar weights and vector aggregate are
\begin{equation}
\begin{aligned}
w_{tin}&=
\frac{\exp(u_{tin}/\tau_v)}
{\sum_{h\in\mathcal S_{tj}}\exp(u_{tih}/\tau_v)},\\
\bar v_{tj}&=\sum_{n\in\mathcal S_{tj}}w_{tin}v_{in},
\end{aligned}
\label{eq:app-group-readout}
\end{equation}
Here $\tau_v>0$ is the token-level softmax temperature,
$w_{tin}\in(0,1)$ is the normalized weight of selected visual position $n$,
and $\sum_{n\in\mathcal S_{tj}}w_{tin}=1$. The aggregate
$\bar v_{tj}\in\mathbb R^d$ is the weighted block-update value for source-time
group $j$ at Planner turn $t$. Define
\begin{equation}
\operatorname{RMS}(x)
=\sqrt{\frac{1}{d}\sum_{h=1}^{d}x_h^2}.
\label{eq:app-rms}
\end{equation}
For $x=(x_1,\ldots,x_d)\in\mathbb R^d$, $x_h$ is its $h$-th coordinate and
$\operatorname{RMS}(x)\in\mathbb R_{\geq0}$ is its root-mean-square magnitude.
This coordinate index $h\in[d]$ is local to Eq.~\eqref{eq:app-rms}.
Before the global bound, the residual from group $G_j$ is
\begin{equation}
\delta_{tj}^{(0)}=
g\,\alpha_{c_i}
\frac{\operatorname{RMS}(q_t)}
{\max(\operatorname{RMS}(\bar v_{tj}),\epsilon)}
\bar v_{tj},
\label{eq:app-unbounded-delta}
\end{equation}
Here $\delta_{tj}^{(0)}\in\mathbb R^d$ is the unbounded group residual;
superscript $(0)$ means ``before the shared global bound.'' The scalar
$g\geq0$ is the global residual-gain multiplier, and $\alpha_{c_i}\geq0$ is
the gain selected by origin role $c_i$. The RMS ratio matches the readout scale
to query $q_t$, while the same $\epsilon=10^{-6}$ used in normalization
protects a zero readout.

\paragraph{Exact receiver-anchor resolution.}
For each source timestamp $\tau_j$, LAVE searches the originating Tool
observation for a line of the form
\texttt{12.3s: ...}. Character offsets map the complete line to Planner tokens;
an exact token-subsequence search is the fallback. The final token of the line
is the receiver position $\pi_{tj}\in[L_t]$. The symbol $\pi_{tj}$ is a
discrete Planner token position and is deliberately distinct from the
source-time probability $p_{tj}$. The main text denotes this receiver anchor by
$p_{tj}$; the appendix uses $\pi_{tj}$ only to disambiguate the anchor from the
probability, without changing the operation. Before injection, the selected tokens are
decoded and the timestamp is parsed again. A group is skipped when the line is
missing, the expected timestamp is absent from the decoded span, or the token
position falls outside the current prompt. This strict contract prevents a
visual update from being written to a semantically unrelated line.
Let $\mathcal J_t^{\mathrm{res}}\subseteq\mathcal J_t^+$ denote exactly the
group indices that pass these checks and therefore have a defined anchor
$\pi_{tj}$.

The line-final token is used because it has causally read both the timestamp
and the complete visible description on that line. Injecting after receiver
block $\ell_r$ leaves the text and tokenization unchanged, while all later
blocks can integrate the visible observation and its aligned latent residual.

\paragraph{Accumulation and the explicit Bound operator.}
Multiple groups may resolve to the same line-final token. Their unbounded
updates are first accumulated by receiver position:
\begin{equation}
D_{t,p}=\sum_{\substack{j\in\mathcal J_t^{\mathrm{res}}\\\pi_{tj}=p}}
\delta_{tj}^{(0)}.
\label{eq:app-position-delta}
\end{equation}
In Eq.~\eqref{eq:app-position-delta}, $p\in[L_t]$ is a generic Planner prompt
position and $D_{t,p}\in\mathbb R^d$ is the sum of all unbounded group
residuals whose receiver anchor equals $p$. Let
$\mathcal A_t=\{\pi_{tj}:j\in\mathcal J_t^{\mathrm{res}}\}$ be the set of distinct
resolved receiver positions. The combined
update magnitude is
\begin{equation}
R_t=\left(\sum_{p\in\mathcal A_t}
\operatorname{RMS}(D_{t,p})^2\right)^{1/2}.
\label{eq:app-combined-rms}
\end{equation}
Thus $R_t\in\mathbb R_{\geq0}$ is the Euclidean aggregation of per-position RMS
magnitudes, not the RMS of a concatenated prompt state. If
$\mathcal A_t=\varnothing$, LAVE defines $R_t=0$ and performs no write.
The shared bound scale, bounded group residual, and position-wise update are
\begin{equation}
\begin{aligned}
\gamma_t&=\min\!\left(
1,\frac{\rho_{\max}\operatorname{RMS}(q_t)}
{\max(R_t,\epsilon)}\right),\\
\delta_{tj}&=\gamma_t\delta_{tj}^{(0)},
\qquad j\in\mathcal J_t^{\mathrm{res}},\\
\Delta_{t,p}&=
\sum_{\substack{j\in\mathcal J_t^{\mathrm{res}}\\\pi_{tj}=p}}
\delta_{tj}
=\gamma_tD_{t,p}.
\end{aligned}
\label{eq:app-bound-scale}
\end{equation}
Here $\rho_{\max}\geq0$ is the maximum permitted ratio between the combined
latent update and query RMS, $\gamma_t\in[0,1]$ is one shared scale for all
receiver positions at turn $t$, $\delta_{tj}\in\mathbb R^d$ is the bounded
group residual denoted by the same symbol in the main text, and
$\Delta_{t,p}\in\mathbb R^d$ is the sum written at position $p$. If
$\operatorname{RMS}(q_t)=0$, the definition gives
$\gamma_t=0$ and hence $\Delta_{t,p}=0$ for every position. For all query
states, it follows directly that
\begin{equation}
\left(\sum_{p\in\mathcal A_t}
\operatorname{RMS}(\Delta_{t,p})^2\right)^{1/2}
\leq\rho_{\max}\operatorname{RMS}(q_t).
\label{eq:app-bound-guarantee}
\end{equation}
When $\operatorname{RMS}(q_t)>0$, dividing both sides recovers the ratio form
used to interpret $\rho_{\max}$; the undivided form above also remains defined
for a zero query.
This shared multiplication is the explicit implementation of
$\operatorname{Bound}(\cdot)$ in the main text: it maps each raw
$\delta_{tj}^{(0)}$ to the bounded $\delta_{tj}$ while limiting their combined
update across all receiver positions. It therefore preserves relative group
magnitudes instead of clipping every group independently.

Finally, the receiver state is
\begin{equation}
\widetilde H_{t,\mathrm{out}}^{(\ell_r)}[p]
=H_{t,\mathrm{out}}^{(\ell_r)}[p]
+\sum_{\substack{j\in\mathcal J_t^{\mathrm{res}}\\\pi_{tj}=p}}
\delta_{tj}
=H_{t,\mathrm{out}}^{(\ell_r)}[p]+\Delta_{t,p}.
\label{eq:app-final-injection}
\end{equation}
Here $H_{t,\mathrm{out}}^{(\ell_r)}[p]\in\mathbb R^d$ is the original Planner
state immediately after receiver block $\ell_r$, and
$\widetilde H_{t,\mathrm{out}}^{(\ell_r)}[p]$ is the modified state passed to
block $\ell_r+1$. Positions outside $\mathcal A_t$ are unchanged, which is
equivalent to setting $\Delta_{t,p}=0$ there.
LAVE applies this operation only to planning prefill. Required Tool-choice
formatting and final-answer generation receive no direct latent update; the
answer can change only through the preceding latent-influenced planning and
Tool trajectory.

\subsection{Experimental Settings and System Details}
\label{app:experimental-settings}

This subsection defines the evaluation and controlled-ablation protocol, then
reports the fixed LAVE, runtime, and backbone configuration.

\subsubsection{Evaluation and Ablation Protocol}
\label{app:experiment-protocol}

\paragraph{Benchmarks and metrics.}
We follow the splits named in the main text: the full Video-MME benchmark, the
LongVideoBench validation set, and the CG-Bench mini set. Subtitles are disabled
for every method. Let $N\in\mathbb N$ be the number of evaluated questions and
$z\in[N]$ index one question. Multiple-choice accuracy is
\begin{equation}
\operatorname{Acc}=\frac{100}{N}
\sum_{z=1}^{N}\mathbf 1[\widehat y_z=y_z].
\label{eq:app-accuracy}
\end{equation}
Here $y_z$ is the ground-truth option label, $\widehat y_z$ is the predicted
option label, and $\mathbf 1[\cdot]$ is one when its Boolean argument is true
and zero otherwise; the factor $100$ reports a percentage. Let
$F_z\in\mathbb N_0$ be the number of source frames sampled by Overview, Skim,
and Focus for question $z$, where $\mathbb N_0=\{0,1,2,\ldots\}$. The reported
efficiency metric is
\begin{equation}
\operatorname{Frames/QA}=\frac{1}{N}\sum_{z=1}^{N}F_z.
\label{eq:app-efficiency-metrics}
\end{equation}
A rendered Overview montage does not collapse its component source frames into
one counted frame; every source frame contributes one to $F_z$.

\paragraph{Matched no-latent reference.}
For each backbone, LAVE and no-latent use the same ordered question identifiers,
benchmark files, model checkpoint and processor, prompt and Tool schemas,
VideoSeek Overview--Skim--Focus loop, frame sampler and timestamp format,
maximum steps, stage-specific generation ceilings, temperature, seed, subtitle
setting, final-answer path, and output schema. The no-latent condition uses the
same visible-channel format, Tool set, observation protocol, and budgets. It
routes no stored latent evidence into planning and receives no alternative
prompt or search budget.
LAVE's hooks read the existing Tool prefill, so they add neither frame replay
nor a second visual encoding pass. Memory is question-scoped, injection is
planning-only, and final answering is text-only in both conditions; Frames/QA
therefore follows the same source-frame counting rule.

\paragraph{Controlled ablations.}
The communication-block study sets $\ell=\ell_r\in\{15,19,23,27\}$ and disables
entropy-constrained frame-time routing, using flat positive-utility token
retrieval in every block condition. Capture and writing use the same candidate
block in each run; the visual-position mask, key/value definition, evidence
budget, Tool-role gains, residual bound, prompts, and decoding policy remain
fixed. The Gaussian control preserves keys, provenance, tensor shapes, and
receiver locations, but replaces each value with a deterministically seeded
zero-mean Gaussian vector rescaled row-wise to the original RMS. This control
tests whether structured value directions outperform norm-matched random
perturbations.

The hyperparameter study is one-factor-at-a-time. Starting from Block 19,
$B=8$, $g=1.0$, and entropy-adaptive coverage, it varies only one factor. The
token-budget sweep fixes $g=1.0$ and entropy-adaptive coverage; the gain sweep
fixes $B=8$ and entropy-adaptive coverage; and the coverage sweep fixes $B=8$
and $g=1.0$. All other routing, injection, question-set, and decoding settings
remain fixed. The evaluated values are
\begin{equation}
\begin{aligned}
B&\in\{4,8,16\},\\
g&\in\{0.5,1.0,2.0\},\\
K&\in\{2,4,8\}\quad\text{or entropy-adaptive}.
\end{aligned}
\label{eq:app-ablation-grid}
\end{equation}
In this grid, $B$ is the integer global evidence-token budget per planning
turn, $g$ is the nonnegative residual-gain multiplier from
Eq.~\eqref{eq:app-unbounded-delta}, and $K$ is a fixed integer number of
source-time groups. ``Entropy-adaptive'' means that $K$ is replaced by
$\widetilde K_t$ from Eq.~\eqref{eq:app-entropy-solution}.
In the fixed-$K$ controls, the same group scores and probabilities are retained
for ranking and diagnostics; only the entropy-derived cardinality is replaced
by $K$. Utility, token-budget allocation, Tool-role gains, receiver block,
timestamp verification, and final-answer policy remain unchanged.

\subsubsection{Fixed Method, Runtime, and Backbone Configuration}
\label{app:runtime-config}

All controlled runs use the same local runtime summarized in
Table~\ref{tab:system-config}. The hardware model is reported without a machine
count or topology assumption.

\begin{center}
\centering
\fontsize{9}{10.2}\selectfont
\setlength{\tabcolsep}{2.2pt}
\begin{tabular}{>{\raggedright\arraybackslash}p{0.30\columnwidth}
                >{\raggedright\arraybackslash}p{0.65\columnwidth}}
\toprule
Setting & Value \\
\midrule
Hardware & NVIDIA RTX A6000 \\
Conda environment & \texttt{qwen35\_vlm\_latent} \\
Python & 3.12.13 \\
pip & 26.1.2 \\
PT / TV & 2.6.0+cu124 / 0.21.0+cu124 \\
CUDA / cuDNN & 12.4 / 9.1.0 \\
Transformers & 5.13.0 \\
Accelerate & 1.14.0 \\
VL utils / Decord & 0.0.14 / 0.6.0 \\
Inference / prec. & Local Transformers / bfloat16 \\
Backbones & Qwen3.5-9B / GLM-4.6V-Flash-9B \\
Agent / Tools & VideoSeek / Overview--Skim--Focus \\
Frame factor & 2 \\
$\alpha_{\rm agent}$ & 2 \\
Max steps & 8 \\
Context ceiling & 32768 tokens \\
Planner ceiling & 4096 tokens \\
Tool-choice ceiling & 1024 tokens \\
Visual Tool ceiling & 32768 tokens \\
Answer ceiling & 64 tokens \\
Temp / seed & 0 / 42 \\
Subtitles & Disabled \\
Execution & Shared local task queue \\
\bottomrule
\end{tabular}
\captionof{table}{Unified system and agent configuration. Generation ceilings
are stage-specific output limits; the context ceiling is reported separately.
PT: PyTorch; TV: Torchvision; VL: vision--language; prec.: precision; Temp:
temperature.}
\label{tab:system-config}
\end{center}

\begin{center}
\centering
\fontsize{9}{10.2}\selectfont
\setlength{\tabcolsep}{2.2pt}
\begin{tabular}{>{\raggedright\arraybackslash}p{0.45\columnwidth}
                >{\raggedright\arraybackslash}p{0.50\columnwidth}}
\toprule
Component & Main value \\
\midrule
Capture / receiver block & 19 / 19 \\
Key / value & $H_{\rm out}$ / $H_{\rm out}-H_{\rm in}$ \\
Capture stage & Original Tool prefill \\
Candidate positions & All visual positions \\
Memory lifetime & Current question \\
Retrieved evidence $B$ & At most 8 tokens \\
Group score & Mean top-4 utilities \\
Group temperature $\tau_g$ & 0.20 \\
Token temperature $\tau_v$ & 0.20 \\
Redundancy weight $\lambda$ & 0.35 \\
$\alpha_{\rm overview}$ & 0.02 \\
$\alpha_{\rm skim}$ & 0.05 \\
$\alpha_{\rm focus}$ & 0.10 \\
Residual multiplier $g$ & $1.0\times$ \\
Combined bound $\rho_{\max}$ & 0.20 query RMS \\
Temporal coverage & Entropy-adaptive \\
Receiver anchor & Timestamp-line final token \\
Active stage & Planning only \\
\bottomrule
\end{tabular}
\captionof{table}{Fixed latent-routing and residual-injection settings used by
the main configuration.}
\label{tab:routing-config}
\end{center}

\noindent\textbf{Functional roles.}
The aligned capture and receiver blocks place Tool keys and Planner queries in
the same hidden space, while the block output and block-local update serve as
the matching key and transferable value. Restricting capture to all visual
positions in the original Tool prefill excludes generated states without
pre-retrieval pooling, and question-scoped memory prevents cross-example
leakage. The global budget $B$ caps total retrieved tokens, while the mean
top-4 group score summarizes a source time without rewarding patch count.
$\tau_g$ shapes the source-time distribution used by entropy-adaptive coverage;
$\tau_v$ controls within-group token readout, and $\lambda$ discounts evidence
already represented in text. The Overview, Skim, and Focus gains scale coarse,
localization, and verification evidence, respectively. Finally, $g$ controls
overall residual strength, $\rho_{\max}$ bounds the accumulated update, and the
timestamp-line final token preserves source alignment. Injection is active only
during planning, so final answering remains text-only.

\noindent\textbf{Runtime and agent roles.}
The A6000 and pinned Conda, Python, pip, PyTorch, Torchvision, CUDA, and cuDNN
entries fix the hardware and tensor-kernel environment. Transformers and
Accelerate provide local model loading and execution, while qwen-vl-utils and
Decord provide multimodal preprocessing and video decoding. Local bfloat16
inference preserves the activation access required by LAVE at a practical
memory cost. Qwen3.5-9B and GLM-4.6V-Flash-9B are the two controlled
backbones; VideoSeek supplies the shared agent loop and its Overview, Skim, and
Focus Tools. The frame factor belongs to the baseline, whereas
$\alpha_{\rm agent}$ belongs to VideoSeek, not to the latent Tool-role gain
$\alpha_{c_i}$ in Eq.~\eqref{eq:app-unbounded-delta}. The step limit bounds
Tool interactions, and the context ceiling bounds the complete rendered
sequence. Temperature zero and seed 42 make decoding deterministic, while
disabling subtitles keeps the comparison visual-only.

\twocolumn[{
\begin{center}
  \centering
  \includegraphics[width=\textwidth]{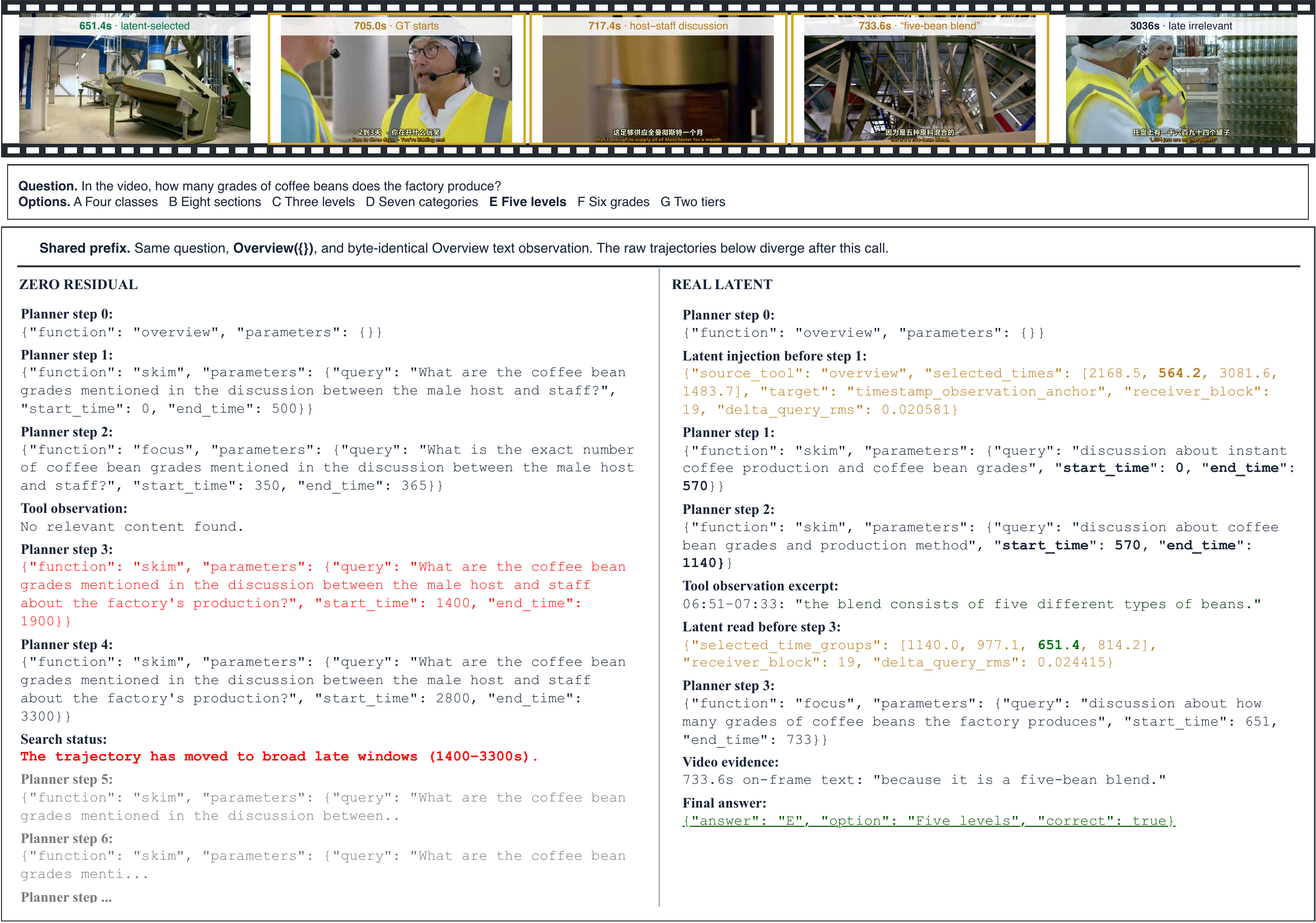}
  \captionof{figure}{Paired trajectory comparison for the coffee-bean question.
  Both conditions share the question and initial Overview Tool call, which
  returns a byte-identical observation. The zero-residual control drifts from a
  failed Focus to late windows. LAVE retains the 651.4\,s source-time group and
  searches the official clue interval. The full trajectories end in F and E,
  respectively.}
  \label{fig:app-paired-case}
\end{center}
\vspace{0.35em}
}]

\begin{center}
\begin{minipage}{0.96\columnwidth}
\centering
\fontsize{9}{10.2}\selectfont
\setlength{\tabcolsep}{2.2pt}
\begin{tabular}{>{\raggedright\arraybackslash}p{0.42\columnwidth}
                >{\raggedright\arraybackslash}p{0.53\columnwidth}}
\toprule
Setting & Value \\
\midrule
Architecture & Cond.-gen. GLM \\
Text precision & bfloat16 \\
Decoder blocks & 40 \\
Hidden / intermediate size & 4096 / 13696 \\
Attention / KV heads & 32 / 2 \\
Max positions & 131072 \\
Vision encoder depth & 24 \\
Vision hidden / output size & 1536 / 4096 \\
Image / patch size & 336 / 14 \\
Temporal patch size & 2 \\
Spatial merge size & 2 \\
Vision attention heads & 12 \\
Processor & \texttt{Glm46VProcessor} \\
\bottomrule
\end{tabular}
\captionof{table}{GLM-4.6V-Flash-9B backbone configuration. Cond.-gen.:
conditional-generation; KV: key--value. LAVE routing and agent settings match
the Qwen3.5-9B condition.}
\label{tab:glm-config}
\end{minipage}
\end{center}

\noindent\textbf{GLM backbone roles.}
The architecture entry corresponds to
\texttt{Glm4vForConditionalGeneration}, and the processor entry corresponds to
\texttt{Glm46VProcessor}; together they identify the exact model and
input-packing interfaces. The bfloat16 entry matches the reported inference
precision. Forty
decoder blocks make Block 19 a valid shared capture/receiver location; the
4096 hidden width fixes the dimension of every LAVE key, value, query, and
residual, while the 13696 intermediate width is internal to each decoder block.
The 32 attention and two KV heads specify grouped-query attention, and 131072
maximum positions exceed the controlled 32768-token context ceiling. On the
visual side, 24 encoder blocks with 12 attention heads operate at width 1536
and project to the 4096 language width. Image size 336 and patch size 14 define
the spatial patch grid; temporal patch size 2 and spatial merge size 2 define
how visual tokens are compressed before entering the language model.

\noindent\textbf{Stage-budget semantics.}
The 32768-token context ceiling bounds the rendered question and trajectory,
not a shared output allowance. The 4096-token Planner budget supports
multi-step reasoning; the isolated 1024-token Tool-choice budget prevents
malformed action selection from consuming the visual Tool allowance. The
32768-token visual Tool ceiling preserves long timestamped observations, and
the 64-token final-answer ceiling enforces a short option-only response.
Together with the eight-step limit, these values bound trajectory depth and
each stage's generation independently.

\noindent\textbf{Configuration scope.}
Table~\ref{tab:system-config} fixes infrastructure and the shared agent
protocol, Table~\ref{tab:routing-config} fixes the main LAVE mechanism, and
Table~\ref{tab:glm-config} records metadata read from the GLM checkpoint. Only
the factors declared in the controlled-ablation grid are varied. The shared
local task queue affects scheduling only, so it cannot change model state,
decoding, or per-question settings. Before the first question, checkpoint
interfaces, runtime flags, and routing values are checked against Tables
\ref{tab:system-config}--\ref{tab:glm-config}; any mismatch is treated as a
configuration error rather than a method result.

\subsection{Paired Case Study: Latent Evidence Redirects Temporal Search}
\label{app:paired-case}

Figure~\ref{fig:app-paired-case} complements the main-paper example with a
matched zero-residual comparison. The replays share the question, initial
Overview action, and byte-identical observation. Because their generated
step-0 thoughts differ, the pair is not an identical-history causal fork, but a
controlled behavioral comparison after matched visible Tool evidence.

The zero-residual control searches 0--500\,s, issues a failed Focus at
350--365\,s, and then drifts to 1400--3300\,s, missing the official
705--745\,s clue interval and predicting F. With latent updates, the Planner
uses Skim windows of 0--570\,s and 570--1140\,s. The second Skim reports five
bean types, the latent read retains the 651.4\,s group, and the next
Focus(651--733\,s) overlaps the clue interval. This run predicts E correctly.

The Focus response contains no relevant content, so it adds no new answer text.
The preceding Skim contains the clue, and the on-frame subtitle text at
733.6\,s corroborates it. The pair therefore supports changed temporal coverage,
not direct answer injection. This diagnostic replay uses receiver block 19,
matching the main evaluation. It provides process-level evidence only, while
the quantitative conclusions remain based on the matched block-19 study.

\newpage
\subsection{Limitations}
\label{app:limitations}

LAVE leaves the Overview--Skim--Focus loop, prompts, visible observations, and
final-answer interface unchanged. However, it requires intermediate Tool and
Planner activations. It captures visual-token block updates during Tool prefill
and writes position-specific residuals during Planner prefill. Black-box VLM
APIs expose neither capability. LAVE therefore assumes open-weight models or
customizable inference. Closed APIs would require provider-side activation
access or validated distillation.

Exact timestamped observation lines are also required; unresolved anchors are
skipped. Hooks and stored states add unquantified memory and latency overhead.
However, LAVE replays no frames and adds no second visual pass.

\end{document}